\documentclass[conference]{IEEEtran}
\usepackage{cite}
\usepackage{amsmath,amssymb,amsfonts}
\usepackage{algorithmic}
\usepackage{graphicx}
\usepackage{textcomp}
\usepackage{xcolor}
\def\BibTeX{{\rm B\kern-.05em{\sc i\kern-.025em b}\kern-.08em
    T\kern-.1667em\lower.7ex\hbox{E}\kern-.125emX}}
\begin{document}

\title{Targeted Background Removal Creates Interpretable Feature Visualizations}

\makeatletter
\newcommand{\linebreakand}{%
  \end{@IEEEauthorhalign}
  \hfill\mbox{}\par
  \mbox{}\hfill\begin{@IEEEauthorhalign}
}
\makeatother

\author{\IEEEauthorblockN{Ian E. Nielsen, Erik Grundeland, Joseph Snedeker, Ravi P. Ramachandran}
\IEEEauthorblockA{
\textit{Rowan University}\\
Glassboro, NJ USA \\
(nielseni6,grunde72,snedek95)@students.rowan.edu, ravi@rowan.edu}
\and
\IEEEauthorblockN{Ghulam Rasool}
\IEEEauthorblockA{
\textit{H. Lee Moffitt Cancer Center}\\
Tampa, FL USA \\
Ghulam.Rasool@moffitt.org}
}

\maketitle

\begin{abstract}
Feature visualization is used to visualize learned features for black box machine learning models. Our approach explores an altered training process to improve interpretability of the visualizations. We argue that by using background removal techniques as a form of robust training, a network is forced to learn more human recognizable features, namely, by focusing on the main object of interest without any distractions from the background. Four different training methods were used to verify this hypothesis. The first used unmodified pictures. The second used a black background. The third utilized Gaussian noise as the background. The fourth approach employed a mix of background removed images and unmodified images. The feature visualization results show that the background removed images reveal a significant improvement over the baseline model. These new results displayed easily recognizable features from their respective classes, unlike the model trained on unmodified data.

\end{abstract}


\section{Introduction}

Deep neural networks are an incredible emerging technology with great potential for continued innovation and application to various fields. In the realm of image classification, neural networks serve as an invaluable tool, demonstrating remarkable accuracy in classifying diverse datasets. These models tend to be opaque, since the only parts which can be directly interpreted by humans are the inputs and the outputs. We say that these models are a black-box. However, through a technique called feature visualization \cite{olah2017feature, nguyen2019understanding, minh2022explainable}, the inside of these networks can be explained. Feature visualization creates a visual representation of key features that the neural network has learned to recognize, providing key insights into how these models learn. Through this process, an image is generated specifically to maximize the activation of one or multiple neurons in the model. The target class output is typically used as the neuron to maximize, resulting in an image that captures the salient features learned by the model for that specific class. 

Feature visualization is considered to be within the field of Explainable AI (XAI). The field of XAI seeks to give a user more trust in the model by creating a visual explanation \cite{nielsen2022robust}. Feature visualization and other methods within this field have uses for medical applications, self-driving cars and other mission-critical applications \cite{das2020opportunities, nielsen2022robust, madhav2022explainable}. 

An example of a feature visualization map can be seen in Figure \ref{fig:sample_FV}. The left side depicts images that the model recognizes as a bird with high certainty. The image to the right is the feature visualization image generated for the bird class. This explainability method incorporates key features learned at multiple layers to create an image which highly activates the target class neuron for bird. From this image, it can be seen that this model recognizes the features of eyes, beaks, and feathers as a bird. 
   
However, this process needs refinement to create the bird image seen in the figure. If the feature visualization is created only to maximize a target class activation, the result becomes unintelligible. These results often appear as high contrast, random looking noise that contains little to no human recognizable features. This is likely because much of the features learned by non-robustly trained models consist of high frequency noise \cite{ilyas2019adversarial}. An example of this problem can be seen in Figure \ref{fig:unregularized_FV} which depicts an unregularized visualization of three classes from an ImageNet \cite{deng2009imagenet} model. These images, as opposed to the feature visualization in Figure \ref{fig:sample_FV}, have no regularizers and are completely unrecognizable, thereby providing no insight into what features the model has come to learn.

\begin{figure}[h]
    \centering
    \includegraphics[width=1.0\columnwidth]{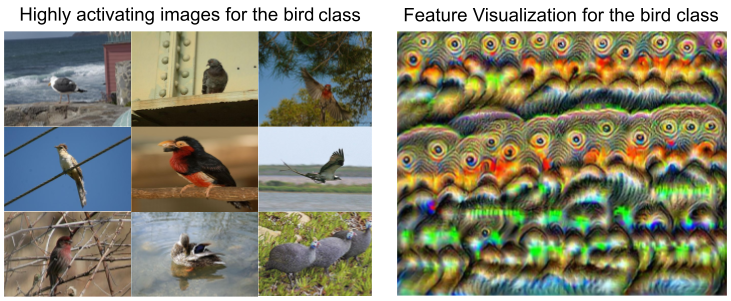}
    \caption{Visual examples of inputs that cause the model to predict bird with a high degree of confidence for ResNet18 \cite{he2016deep_resnet} trained on the PASCAL VOC 2012 dataset \cite{pascal-voc-2012}. The left image shows high activating images for the bird classification. The right shows the feature visualization for the same class.}
    \label{fig:sample_FV}
\end{figure}

In order to create recognizable feature visualization, as in Figure \ref{fig:sample_FV}, regularization algorithms are applied to the image throughout the process of generating the explanation. This approach enhances the visualizations by reducing the effect that noise has during the generation process. The outcome is an explanation that is focused on more recognizable features. Very simple examples of this include shifting the image every few iterations, adding a small amount of Gaussian noise to the image, and penalizing high frequency noise. However, these regularizers are not perfect. They may create a more recognizable image, but can also cause the final image to be a less accurate representation of what the model recognizes as a class. An ideal feature visualization image would be created directly from the model with no modification as to best represent the model’s understanding. 

The goal of this paper is to explore why the unregularized output is so unclear, and how the training process can be used to improve clarity and recognizability of visualization. Multiple sets of training data will be used to assess the impact that our approaches to training have on feature visualizations. The first model will serve as a baseline and will be trained on unmodified data. The other models will be trained on a modified dataset with backgrounds removed, encouraging the model to learn in a more object-focused manner.  Models typically train on a full image which includes other contextual information, and can make connections to parts of the image that are not the object of interest. This means that models often rely on contextual information in the background, unlike humans who can still recognize objects in various contexts. This distinction in learning methods could explain the unrecognizability of feature visualizations to humans. By training the models on a background-removed dataset, we aim to encourage more human-like object recognition and generate more interpretable feature visualizations. We believe that this approach enhances model explainability without distorting features with regularizers.
   
\begin{figure}[h]
    \centering
    \includegraphics[width=1.0\columnwidth]{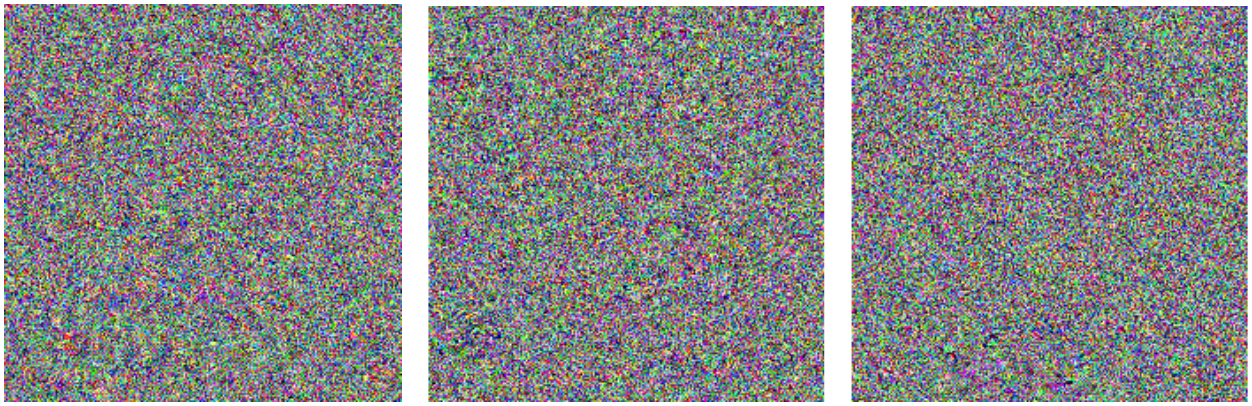}
    \caption{Unregularized feature visualization outputs for three different target classes (cat, person, and potted plant from left to right respectively) from ResNet \cite{he2016deep_resnet} trained on Tiny-ImageNet \cite{le2015tiny_imagenet}. It is clear that these unregularized visualizations do not contain human recognizable features.} % Labels (left to right): goldfish, European fire salamander, school bus
    \label{fig:unregularized_FV}
\end{figure}

\section{Prior Work}

Most of the research around feature visualization seeks to improve explanations by modifying the process of generating them \cite{nguyen2019understanding, olah2017feature}, regardless of how the model is trained. Very little work concentrates on changing the characteristics of the training dataset to create more robust visualizations. Robust models are more resistant to small input perturbations (including noise and adversarial attacks) than their non-robust counterparts, often at the expense of a small decrease in overall accuracy \cite{tsipras2019robustness}. It has been shown that robust models produce more human interpretable explanations than their non-robust counterparts \cite{nielsen2022robust, alvarez2018robustness, bansal2020sam}. In \cite{engstrom2019adversarial}, researchers use adversarial robustness as a learned prior to aid the model in defense against these examples, with minimal accuracy loss. Another common robust model technique is to utilize input filtering. This could consist of monitoring unusual neuron activation patterns that could be from an adversarial example, or utilizing saliency maps as an adversarial filter \cite{ye2020detection}. The model classifies the input image, then applies a saliency map to the image and classifies it again. If the classifications do not match, the input is  likely an adversarial example. The previous research has the goal of making the outputs more understandable to the average person. This paper shares that goal, but approaches it differently. Our method changes the inputs by removing potentially irrelevant information from the background, but keeping all information within the object of interest unchanged. We explore the effect of the training images on the quality of the model's ability to recognize features.

\section{Impacts}

We would like to emphasize the importance of user trust for these models, which are used more widely every day, especially for mission-critical tasks. Allowing the user to visualize and explain learned features can increase that trust significantly. Feature visualization maps have usefulness for both developers who are trying to improve and evaluate their models, as well as users who want to determine whether they can trust the model outputs. This investigation has wide impact and practical applications. For example, feature visualization can assist a machine learning algorithm to show patterns in genomes or bacteria in a way that a person can understand \cite{novakovsky2023obtaining}. This not only improves user trust and understandability immediately, but also helps to understand where a model goes wrong so that it can be fixed. For that reason, we know that this work has the impact of safer implementations for artificially intelligent systems. Creating a more understandable model helps prevent confusion in any real world applications. A model needs to be able to accurately recognize objects in important situations, like self-driving cars, as otherwise there is a danger to the passengers. By improving the visualizations used to explain these models, we are enabling more robust testing and detection of bugs or errors that would otherwise go undetected.

%% Perhaps we should add bullet points detailing the novel contributions

%% We could also add a paragraph detailing the setup of the rest of the paper
%
\section{Techniques and Approaches} 

For our experiments, we train four ResNet18 models on different modified versions of the data. These models all have the same architecture and number of layers, so that we can make a valid comparison. We use the PASCAL VOC 2012 dataset \cite{pascal-voc-2012}, which contains 20 classes. Although the resolution is not uniform across the images, there is a maximum resolution of 500x500. This dataset was chosen over other commonly used datasets, like ImageNet or CIFAR10, because it includes a segmentation map for each image. This can be seen in Figure \ref{fig:trainset_images}. The first column shows the unmodified image, and the second column shows the segmentation map for that image. It provides an exact mask for isolating the foreground and background, which is perfect for creating a dataset with the background accurately removed. 

\begin{figure}[ht]
    \centering

    \includegraphics[width=1.0\columnwidth]{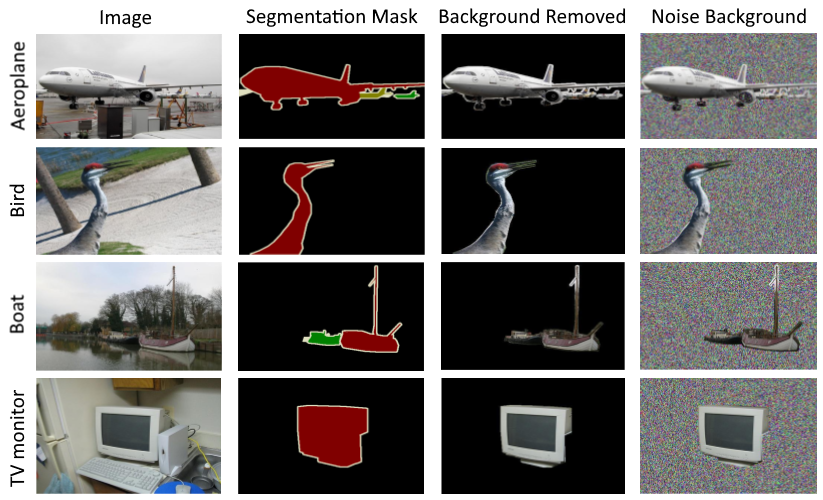}
    \caption{Images from the PASCAL VOC 2012 dataset and the corresponding segmentation maps were used to decide which pixels to remove and replace. The rows are labeled with the class that each image belongs to. From left to right, the images depicted in columns 1, 3 and 4 were used to train the standard, background removed and noise background models respectively.
    }
    \label{fig:trainset_images}
\end{figure}

Using this dataset and its segmentation maps, we created custom training sets for training each model. As mentioned before, the first model is trained on clean images, as seen in column 1 of Figure \ref{fig:trainset_images}. The second model is trained on background removed images. These images contain only the object of interest, with the rest of the image set to zero, as can be seen in column 3 of Figure \ref{fig:trainset_images}. This training set forces this model to only learn the information of the main object in each image, and totally ignore any distractions from the background. The third model is trained with the background replaced with random Gaussian noise, as seen in column 4 of Figure \ref{fig:trainset_images}. This training set determines how the model reacts when the background context is removed, but is still non-uniform. The model will not be able to learn to associate black with background due to the background being random. The fourth and final model is trained on a mixed training set. Half of the dataset is background removed and set to zero, while the other half consists of unmodified images. This training set will show how the model reacts with both background and no background. 

To remove the unwanted background pixels, we performed an elementwise multiplication between the image and the segmentation map. This resulted in all background pixels becoming zero, and all foreground pixels remaining constant. For the noise background, we then added Gaussian noise to only the background pixels. All backgrounds were removed prior to training, so the impact on performance was negligible. 

The Resnet18 network was chosen for this experiment due to its relative simplicity and effectiveness at image classification tasks. While this model was created with ImageNet in mind, the similar resolutions of ImageNet and PASCAL VOC 2012 allow the model to work well with these custom datasets. The parameters for each model were kept the same across all training sets, and each model was trained for 100 epochs. The same test set and validation set was also used for each model, containing normal images similar to those used for training the first model. 

The goal for this paper is to improve upon feature visualizations by focusing on model training rather than by modifying the process of generating visualizations. Many methods exist for generating feature visualizations \cite{nguyen2019understanding} \cite{simonyan2014deep} \cite{nguyen2016multifaceted} \cite{mahendran2016visualizing} \cite{wei2015understanding} \cite{yosinski2015understanding}, but we decided to focus on one of the most basic methods, including activation maximization with L2 regularization as described in \cite{simonyan2014deep}, \cite{nguyen2019understanding} and \cite{olah2017feature}.

\section{Results}

All four models were trained up to 100 epochs utilizing the ResNet18 architecture. All hyperparameters were set the same for all models. The final accuracy scores and losses can be seen below in Table 1. Each model reached at least 99 percent training accuracy, with relatively low losses. The model trained on standard data received the lowest accuracy score, though there are no significant differences between any of the training accuracy scores. The validation accuracy is lower on all the background removed models, but this is to be expected since robust models tend to have lower validation accuracy than their non-robust counterparts \cite{tsipras2019robustness}, but they generalize better for out of distribution data.

\begin{table}[h!]
\caption{Training accuracy for the ResNet18 models trained on the PASCAL VOC 2012 dataset.}
\label{table:1}
\centering
\begin{tabular}{|c | c | c | c|}
 \hline
 \textbf{Model Trained On} & \textbf{Train Accuracy} & \textbf{Val Accuracy} & \textbf{Loss} \\ [0.5ex]
  \hline\hline
 Standard Dataset & 99.49\% & 94.81\% & 0.022 \\ \hline
 Black Background & 99.62\% & 83.55\% & 0.002 \\ 
 Dataset &  &  &  \\ \hline
 Noise Background & 99.69\% & 78.95\% & 0.008 \\ 
 Dataset &  &  &  \\ \hline
 Mixed Dataset & 99.95\% & 88.84\% & 0.003 \\
 
 \hline
\end{tabular}

\end{table}

With the models successfully trained, the results of each feature visualization could be generated. An equal L2 regularizer \cite{olah2017feature} was applied to each output image to enhance the images enough to make comparisons between them. The results of this feature visualization can be seen in Figure \ref{fig:results}.

\begin{figure}[h]
    \centering
    \includegraphics[width=1.0\columnwidth]{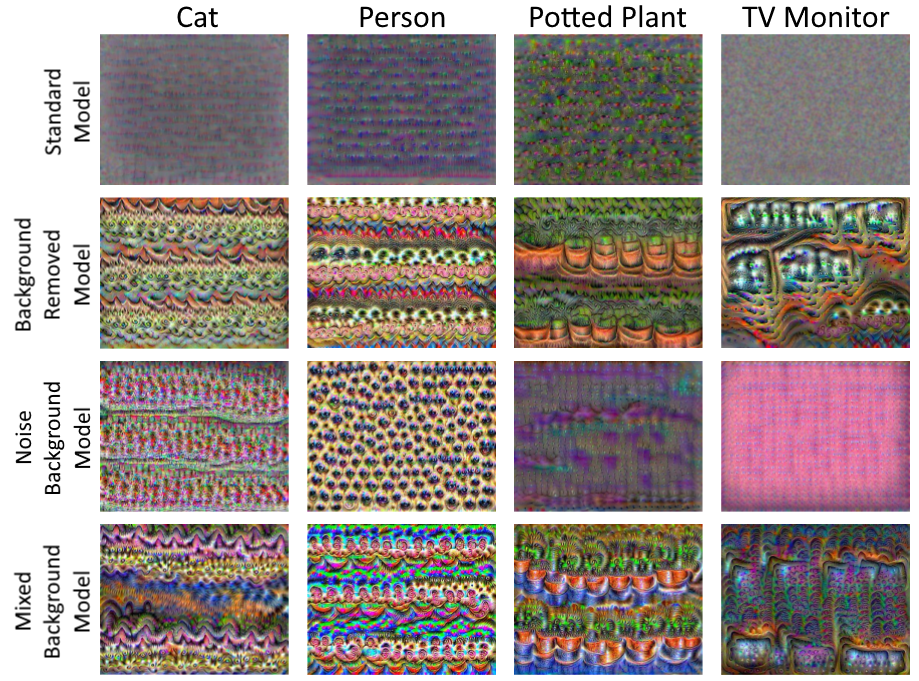}
    \caption{Feature visualization results for four image classes for each of the trained ResNet models. Each row depicts visualizations generated by the models trained on standard data and on different sets of background removed data. The columns are labeled with the class that each image belongs to. }
    \label{fig:results}
\end{figure}

These images show clear differences between each of the trained models. The first model was trained on unmodified PASCAL VOC 2012 images. Here, the problem that this paper addresses can be seen clearly. In all feature visualization results for the standard model, no obviously distinguishable features are seen. The images consist mainly of noise and provide little to no information about what the model has learned for these classes. Of the standard model results, the potted plant class shows features most expected of this label, and only contains flecks of green. 

The results from the model trained on background removed images show vast improvement. For all the classes seen in Figure \ref{fig:results}, the background removed model clearly shows the features it has learned to associate with each class. For the cat label, it can be seen that the model learned to recognize cats through the ears and eyes. For the potted plant label, the model has learned to recognize an orange pot and green leaves. From these observations, it is clear that training the model without the background has led to more human recognizable features being present in the explanations. 

The third model trained used random noise as its background instead of zeros. This has led to significant changes in the results. As seen in Figure \ref{fig:results}, this model has not been as successful as the background removed model. For the background removed model, the input is the object surrounded by all zeros. The edges are easy to define, and the object is always clear. However, with the background set to random values, the edges become harder for the network to differentiate and the network has a more difficult time determining what is important to look at. This model shows only a small improvement over the standard model and creates less human recognizable features. For the cats label, the noise model shows sharper defined edges than the standard model, but none of the images are recognizable as cats. Other classes result in only high frequency noise, similar to the noise used for the backgrounds of the training images, as is seen in the TV monitor label. 

Finally, the mixed dataset model also shows improvement over the standard model. This dataset was trained half on background removed images and half on standard images. Through the feature visualization figures, this training method reveals that the images still clearly show the main focus, but also include some details that were not found within the background removed model. In the cats label for example, the background removed model contains features specifically pertaining to the cat alone. In contrast, the mixed model depicts some of the same features, but also includes extra details that surround the cats (perhaps relating to what they are sitting on). This issue can be seen again for the potted plants class as well. The background removed model shows only the pots and plants. However, the mixed background model shows not only the potted plant, but also parts of a table or platform the plant rests on, thereby adding some context to the image.

\section{Discussion and Future Work}

This paper covers a proof of concept that removing the background of the training set does significantly increase the feature visualization quality. However, this method requires accurately separating the foreground from the background of each training image. The PASCAL VOC 2012 dataset already had segmentation maps for each image, which is why it was used for this experiment. However, in other applications, previously generated segmentation masks may not be readily available. 

In the future, state-of-the-art segmentation networks could be used to remove the background without any need for human generated segmentation maps. Currently, we have not been able to find a model which efficiently and reliably selects only background pixels from images across all the datasets we tested. Automatic removal could become achievable with a more accurate model for the segmentation of background pixels. One challenge will be that the models must be trained on manually segmented data that may not generalize to all other datasets. 

% One notable model is U$^2$-Net \cite{qin2020u2}, which is a likely contender to use in future research.
%
\section{Conclusion}

We argue that in order to improve model explanations, we should work on improving how the model learns, rather than only focusing on improving the methods used to create the visualizations. We see how effective this can be in our results. In this work, we showed that by selectively choosing what the network sees during training, we can create models which learn more human recognizable and interpretable features. We found that the black background and mixed datasets produced the most interpretable visualizations. The noise background dataset also produced much more interpretable explanations than the standard data, but not nearly as well as the other two datasets. Our training approaches also improved overall accuracy of each model, with the mixed dataset performing the best. These results were achieved in an ideal scenario using the PASCAL VOC 2012 dataset, where we were able to remove the background at the exact edge of the object of interest before training. We discuss our attempt to remove the background using attribution, but this approach did not accurately segment the images. However, this approach can be scaled up by using a state-of-the-art segmentation network to remove the background. We show that our results are more interpretable to humans than the vanilla model. We also believe that these approaches should be evaluated with quantitative metrics to confirm our conclusions. This is beyond the scope of this paper, as there is no current consensus on how best to perform quantitative evaluations of feature visualizations.

\section{Acknowledgments}
This work was supported by the (1) NSF Awards ECCS-1903466, OAC-2008690, OAC-2234836 and (2) US Department of Education GAANN program Award Number P200A180055.

\bibliographystyle{IEEEtran}
\bibliography{main}

\begin{thebibliography}{10}
\providecommand{\url}[1]{#1}
\csname url@samestyle\endcsname
\providecommand{\newblock}{\relax}
\providecommand{\bibinfo}[2]{#2}
\providecommand{\BIBentrySTDinterwordspacing}{\spaceskip=0pt\relax}
\providecommand{\BIBentryALTinterwordstretchfactor}{4}
\providecommand{\BIBentryALTinterwordspacing}{\spaceskip=\fontdimen2\font plus
\BIBentryALTinterwordstretchfactor\fontdimen3\font minus
  \fontdimen4\font\relax}
\providecommand{\BIBforeignlanguage}[2]{{%
\expandafter\ifx\csname l@#1\endcsname\relax
\typeout{** WARNING: IEEEtran.bst: No hyphenation pattern has been}%
\typeout{** loaded for the language `#1'. Using the pattern for}%
\typeout{** the default language instead.}%
\else
\language=\csname l@#1\endcsname
\fi
#2}}
\providecommand{\BIBdecl}{\relax}
\BIBdecl

\bibitem{olah2017feature}
C.~Olah, A.~Mordvintsev, and L.~Schubert, ``Feature visualization,''
  \emph{Distill}, vol.~2, no.~11, p.~e7, 2017.

\bibitem{nguyen2019understanding}
A.~Nguyen, J.~Yosinski, and J.~Clune, ``Understanding neural networks via
  feature visualization: A survey,'' in \emph{Explainable AI: interpreting,
  explaining and visualizing deep learning}.\hskip 1em plus 0.5em minus
  0.4em\relax Springer, 2019, pp. 55--76.

\bibitem{minh2022explainable}
D.~Minh, H.~X. Wang, Y.~F. Li, and T.~N. Nguyen, ``Explainable artificial
  intelligence: a comprehensive review,'' \emph{Artificial Intelligence
  Review}, pp. 1--66, 2022.

\bibitem{nielsen2022robust}
I.~E. Nielsen, D.~Dera, G.~Rasool, R.~P. Ramachandran, and N.~C. Bouaynaya,
  ``{Robust Explainability}: A tutorial on gradient-based attribution methods
  for deep neural networks,'' \emph{IEEE Signal Processing Magazine}, vol.~39,
  no.~4, pp. 73--84, 2022.

\bibitem{das2020opportunities}
A.~Das and P.~Rad, ``Opportunities and challenges in explainable artificial
  intelligence (xai): A survey,'' \emph{arXiv preprint arXiv:2006.11371}, 2020.

\bibitem{madhav2022explainable}
A.~S. Madhav and A.~K. Tyagi, ``Explainable artificial intelligence (xai):
  connecting artificial decision-making and human trust in autonomous
  vehicles,'' in \emph{Proceedings of Third International Conference on
  Computing, Communications, and Cyber-Security: IC4S 2021}.\hskip 1em plus
  0.5em minus 0.4em\relax Springer, 2022, pp. 123--136.

\bibitem{ilyas2019adversarial}
A.~Ilyas, S.~Santurkar, D.~Tsipras, L.~Engstrom, B.~Tran, and A.~Madry,
  ``Adversarial examples are not bugs, they are features,'' \emph{Advances in
  neural information processing systems}, vol.~32, 2019.

\bibitem{deng2009imagenet}
J.~Deng, W.~Dong, R.~Socher, L.-J. Li, K.~Li, and L.~Fei-Fei, ``Imagenet: A
  large-scale hierarchical image database,'' in \emph{2009 IEEE conference on
  computer vision and pattern recognition}.\hskip 1em plus 0.5em minus
  0.4em\relax Ieee, 2009, pp. 248--255.

\bibitem{he2016deep_resnet}
K.~He, X.~Zhang, S.~Ren, and J.~Sun, ``Deep residual learning for image
  recognition,'' in \emph{Proceedings of the IEEE conference on computer vision
  and pattern recognition}, 2016, pp. 770--778.

\bibitem{pascal-voc-2012}
M.~Everingham, L.~Van~Gool, C.~K.~I. Williams, J.~Winn, and A.~Zisserman, ``The
  {PASCAL} {V}isual {O}bject {C}lasses {C}hallenge 2012 {(VOC2012)}
  {R}esults,''
  http://www.pascal-network.org/challenges/VOC/voc2012/workshop/index.html.

\bibitem{le2015tiny_imagenet}
Y.~Le and X.~Yang, ``Tiny imagenet visual recognition challenge,'' \emph{CS
  231N}, vol.~7, no.~7, p.~3, 2015.

\bibitem{tsipras2019robustness}
D.~Tsipras, S.~Santurkar, L.~Engstrom, A.~Turner, and A.~Madry, ``Robustness
  may be at odds with accuracy,'' in \emph{International Conference on Learning
  Representations}, no. 2019, 2019.

\bibitem{alvarez2018robustness}
D.~Alvarez-Melis and T.~S. Jaakkola, ``On the robustness of interpretability
  methods,'' \emph{arXiv preprint arXiv:1806.08049}, 2018.

\bibitem{bansal2020sam}
N.~Bansal, C.~Agarwal, and A.~Nguyen, ``Sam: The sensitivity of attribution
  methods to hyperparameters,'' in \emph{Proceedings of the ieee/cvf conference
  on computer vision and pattern recognition}, 2020, pp. 8673--8683.

\bibitem{engstrom2019adversarial}
L.~Engstrom, A.~Ilyas, S.~Santurkar, D.~Tsipras, B.~Tran, and A.~Madry,
  ``Adversarial robustness as a prior for learned representations,''
  \emph{arXiv preprint arXiv:1906.00945}, 2019.

\bibitem{ye2020detection}
D.~Ye, C.~Chen, C.~Liu, H.~Wang, and S.~Jiang, ``Detection defense against
  adversarial attacks with saliency map,'' \emph{International Journal of
  Intelligent Systems}, 2020.

\bibitem{novakovsky2023obtaining}
G.~Novakovsky, N.~Dexter, M.~W. Libbrecht, W.~W. Wasserman, and S.~Mostafavi,
  ``Obtaining genetics insights from deep learning via explainable artificial
  intelligence,'' \emph{Nature Reviews Genetics}, vol.~24, no.~2, pp. 125--137,
  2023.

\bibitem{simonyan2014deep}
K.~Simonyan, A.~Vedaldi, and A.~Zisserman, ``Deep inside convolutional
  networks: Visualising image classification models and saliency maps,'' in
  \emph{In Workshop at International Conference on Learning
  Representations}.\hskip 1em plus 0.5em minus 0.4em\relax Citeseer, 2014.

\bibitem{nguyen2016multifaceted}
A.~Nguyen, J.~Yosinski, and J.~Clune, ``Multifaceted feature visualization:
  Uncovering the different types of features learned by each neuron in deep
  neural networks,'' \emph{arXiv preprint arXiv:1602.03616}, 2016.

\bibitem{mahendran2016visualizing}
A.~Mahendran and A.~Vedaldi, ``Visualizing deep convolutional neural networks
  using natural pre-images,'' \emph{International Journal of Computer Vision},
  vol. 120, no.~3, pp. 233--255, 2016.

\bibitem{wei2015understanding}
D.~Wei, B.~Zhou, A.~Torrabla, and W.~Freeman, ``Understanding intra-class
  knowledge inside cnn,'' \emph{arXiv preprint arXiv:1507.02379}, 2015.

\bibitem{yosinski2015understanding}
J.~Yosinski, J.~Clune, A.~Nguyen, T.~Fuchs, and H.~Lipson, ``Understanding
  neural networks through deep visualization,'' \emph{arXiv preprint
  arXiv:1506.06579}, 2015.

\end{thebibliography}

\end{document}